\documentclass[a4paper,fleqn]{cas-sc}

\usepackage[numbers]{natbib}

\def\tsc#1{\csdef{#1}{\textsc{\lowercase{#1}}\xspace}}
\tsc{WGM}
\tsc{QE}
\tsc{EP}
\tsc{PMS}
\tsc{BEC}
\tsc{DE}

\begin{document}
\let\WriteBookmarks\relax
\def\floatpagepagefraction{1}
\def\textpagefraction{.001}
\shorttitle{Multi-objective Reinforcement Learning With Augmented States Requires Rewards After Deployment}
\shortauthors{P. Vamplew and C. Foale}

\title [mode = title]{Multi-objective Reinforcement Learning With Augmented States Requires Rewards After Deployment}

\author{Peter Vamplew}[
                        orcid=0000-0002-8687-4424]
\cormark[1]
\ead{p.vamplew@federation.edu.au}
\credit{Conceptualization of this study, Writing - Original draft preparation}

\author{Cameron Foale}[
                        orcid=0000-0003-2537-0326]
\ead{c.foale@federation.edu.au}
\credit{Conceptualization of this study, Writing - Original draft preparation}

\affiliation{organization={Institute of Innovation, Science and Sustainability, Federation University Australia},
                addressline={University Drive}, 
                city={Mt Helen},
                postcode={3350}, 
                state={Victoria},
                country={Australia}}

\cortext[cor1]{Corresponding author}

\begin{abstract}
This research note identifies a previously overlooked distinction between multi-objective reinforcement learning (MORL), and more conventional single-objective reinforcement learning (RL). It has previously been noted that the optimal policy for an MORL agent with a non-linear utility function is required to be conditioned on both the current environmental state and on some measure of the previously accrued reward. This is generally implemented by concatenating the observed state of the environment with the discounted sum of previous rewards to create an \emph{augmented state}. While augmented states have been widely-used in the MORL literature, one implication of their use has not previously been reported -- namely that they require the agent to have continued access to the reward signal (or a proxy thereof) after deployment, even if no further learning is required. This note explains why this is the case, and considers the practical repercussions of this requirement.
\end{abstract}




\maketitle

\section{Introduction and Background}

Multi-objective reinforcement learning (MORL) is a generalisation of the standard single-objective reinforcement learning (RL) model, designed for tasks where there are multiple, conflicting objectives \cite{roijers2013survey, hayes2022practical,felten2024multi}. MORL assumes that the environment in which the agent is operating can be represented as a Multi-objective Markov Decision Process (MOMDP). A MOMDP is represented by the tuple $\langle S, A, T, \gamma, \mu, \mathbf{R} \rangle$, where: 
\begin{itemize}
    \item $S$ is the state space and $A$ is the action space
    \item $T \colon S \times A \times S \to \left[ 0, 1 \right]$ is a probabilistic transition function
    \item $\gamma \in [0, 1)$ is a discount factor
    \item $\mu \colon S \to [0,1]$ is a probability distribution over initial states 
    \item $\mathbf{R}: S \times A \times S \to \mathbb{R}^d$ is a vector-valued reward function which specifies the immediate reward for each of the $d$ objectives
\end{itemize}

For single-objective RL, the aim is to discover a policy $\pi$ which maximises some measure of the return from the rewards - usually either the finite-horizon undiscounted reward or the infinite-horizon cumulative discounted reward. These concepts are not directly applicable in the multi-objective case however as a policy which is optimal with regards to one objective may be sub-optimal with regards to another objective. In general there will be a set of policies with returns that are Pareto-optimal, with each representing a differing trade-off between the objectives. In the utility-based paradigm for MORL it is generally assumed that there is a \emph{utility function} $u$, which maps the multi-objective value of a policy to a scalar value \cite{hayes2022practical}. While some MORL methods assume $u$ is a linear function (e.g. \citet{cao2021efficient}), more generally it may be any monotonically-increasing non-linear function (e.g. \citet{parisi2016multi}, \citet{bai2022joint}). For the remainder of this paper we will consider the case where $u$ is non-linear.

As well as defining $u$, a practitioner applying MORL must decide which of two alternative optimisation criteria is to be optimised. These differ in terms of when $u$ is applied.  Scalarised Expected Reward (SER) applies $u$ to the expected vector reward (Equation \ref{eq:ser}), and is appropriate for scenarios where we are concerned about the trade-off in objectives over multiple episodes (i.e. where the policy will be executed multiple times). Expected Scalarised Reward (Equation \ref{eq:esr}) applies $u$ within the expectation operator, and is suited to tasks where we care about the trade-off between objectives achieved within each individual episode \footnote{For a fuller discussion of the SER and ESR distinction and the circumstances to which each is suited, refer to \citet{hayes2022practical}}.

\begin{equation}
    \text{Scalarised Expected Return: }
    V_{u}^{\pi} = u\left(\mathbb{E} \left[ \sum\limits^\infty_{i=0} \gamma^i {\bf r}_i \:\middle|\: \pi, \mu \right]\right)
    \label{eq:ser}
\end{equation}

\begin{equation}
    \text{Expected Scalarised Return: }
    V_{u}^{\pi} = \mathbb{E} \left[ u\left( \sum\limits^\infty_{i=0} \gamma^i {\bf r}_i \right) \:\middle|\: \pi, \mu \right]
    \label{eq:esr}
\end{equation}

\begin{figure}
\centering
\includegraphics[width=.6\textwidth]{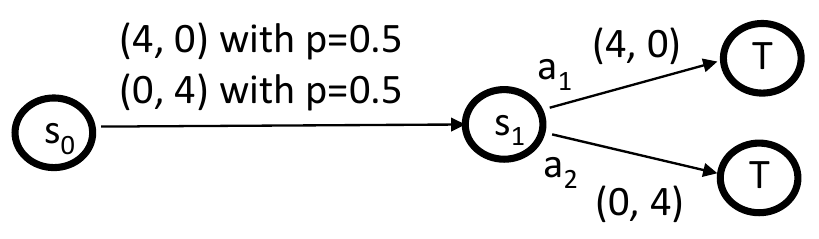}
\caption{\label{fig:2-state-momdp-esr} A simple MOMDP illustrating the need for an augmented state. The optimal choice of action at state $s_1$ depends on the reward received when transiting from state $s_o$ to $s_1$.}
\end{figure}

If the utility function $u$ is non-linear this has significant implications for the nature of the optimal policy. The correct choice of action at a given state may depend not just on that state, but also on the rewards which the agent has received prior to reaching that state. Consider the simple example in Figure \ref{fig:2-state-momdp-esr}. From the starting state $s_0$, executing any action transitions to state $s_1$. The reward received will be $(4,0)$ or $(0, 4)$ with equal probability. From $s_1$ both actions transition to a terminal state but with differing deterministic rewards of $(0, 4)$ for action $a_0$ and $(4, 0)$ for action $a_1$. Assume that the utility function $u$ is designed for fairness amongst objectives by calculating the minimum value across both objectives, and we are optimising for ESR\footnote{We focus on ESR here as it is the simpler case, but the need for augmented states can also apply for SER - see \cite{vamplew2022impact}.}. Under these conditions, the optimal choice of action at $s_1$ depends on the reward received prior to reaching that state. If the agent has received $(4, 0)$ then it should select $a_1$ otherwise it should select $a_2$. This ensures an ESR return of 4 in all episodes. In contrast, any policy (either deterministic or stochastic in nature) conditioned purely on the current state will select an inferior action in $s_1$ half of the time, meaning those episodes will have an ESR return of 0, giving rise to a sub-optimal mean ESR return over all episodes of just 2.

As illustrated by this example, in general in order to achieve optimal behaviour, the agent's policy must be conditioned on both the environmental state and some additional history of the prior rewards. \citet{geibel2006reinforcement} coined the terminology \emph{augmented state} to refer to this concatenation of the environmental state with the summed discounted rewards from the current episode.

\begin{equation}
    \text{Accumulated Reward: }
    r_t^{acc} = \sum\limits^{t-1}_{i=0} \gamma^i {\bf r}_i
    \label{eq:accumulated-reward}
\end{equation}

\begin{equation}
    \text{Augmented State: }
    s_t^A = s_t \cup r_t^{acc}
    \label{eq:augmented-state}
\end{equation}

Some special cases exist where conditioning the policy on the augmented state may not be required, because the value of $r_t^{acc}$ is always the same whenever any specific state $s$ is encountered. Examples include fully-deterministic environments \cite{vamplew2022impact}, or environments where non-zero rewards are only received on steps which end in a terminal state \cite{issabekov2012empirical}. We note that the requirement to condition the policy on the augmented state is independent of the nature of the MORL algorithm, and applies to both value-based and policy-search algorithms. However for value-based algorithms like multi-objective Q-learning, the $r_t^{acc}$ value may be required even in cases like deterministic environments as it is explicitly used within the action-selection process. 

\section{Implications of Augmented States in MORL}

\subsection{Augmented states require post-deployment rewards} While the concept of augmented states has been widely used in MORL research, the majority of this work focuses on the performance of the agent during learning (for example, \citet{reymond2023actor,yu2023fair,chouaki2025fairness,Ding_Vamplew_Foale_Dazeley_2025,pengmulti,peng2025multi,zhao2025adaptive}). As a result, an important practical issue has so far been overlooked. As the policy of the MORL agent is conditioned on the augmented state, the agent must continue to be provided with rewards after it is deployed so that the accumulated reward component of the augmented state can be maintained. This is true even if the agent is simply following a fixed policy and no longer learning.

This forms a further point of distinction between MORL and conventional RL based on scalar rewards. In the latter case, if the agent is deployed within a static environment and is not further updating its policy, then rewards are no longer required.

In some scenarios the provision of post-deployment rewards will not pose any problems. For example, if RL is applied in the context of a video game or a simulator then the training and deployment environments are the same, and so the same reward mechanisms are available post-deployment \citep{wurman2022outracing, li2025comprehensive}. This also holds for other domains which are primarily digital in nature, such as on-line financial trading \citep{avramelou2024deep}.

However in other application domains, particularly those involving a physically-situated agent, providing rewards in the deployment environment may be challenging or overly expensive. For example \citet{Levine-RSS-19} and \citet{ibarz2021train} describe end-to-end RL training on a variety of robotic tasks such as folding clothing, opening doors and manipulating fluids. The robots carrying out these tasks perceive the environment visually, but a variety of different forms of additional instrumentation are required within the environment in order to be able to evaluate the reward function, such as thermal cameras, motion capture sensors and accelerometers. In the context of single-objective RL the agent can be trained with the aid of these sensors, and then deployed without it as the rewards are no longer required. However in the MORL contexts identified earlier where post-deployment rewards are necessary, the need for this additional instrumentation may render these approaches infeasible. Even if this instrumentation is available post-deployment, \citet{parisi2024monitored} note that it may fail. For an MORL agent utilising an augmented state this would make it unable to continue to perform according to its learned policy due to the lack of observable rewards even if the sensors used to observe the environmental state remain active.

Similar issues may arise in the context of sim2real RL. When training within the simulator, additional information about internal variables may be used to determine rewards during training \cite{luo2019end}. However once the learned policy is deployed to a physical system, that additional information is no longer available, and hence neither are the rewards.

\subsection {Reward models as proxies for post-deployment rewards}
Therefore it is important to consider how MORL agents based on an augmented state can be deployed in application domains where providing an ongoing reward signal is problematic. We suggest that the most viable means of addressing this is through the creation of a learned proxy reward model that emulates the MOMDP's reward function $\mathbf{R^P}: S \times A \times S \to \mathbb{R}^d$.

The use of reward models in reinforcement learning is already widespread. \citet{yu2025reward} review a variety of contexts in which the development of a reward model is central to the reinforcement learning process, including inverse reinforcement learning, learning from goals, and learning from preferences. The latter area in particular has seen significant research activity in recent years, driven by the use of reinforcement learning from human feedback (RLHF) as a method for refining generative AI systems \cite{kaufmann2023survey}.

We note that the context in which these models are developed differs substantially from the MORL context which we are considering. They address situations where no ground-truth reward is available even during the training phase, whereas our context assumes that reward is available during training but not in deployment. Hence most prior work on reward models involves the relatively difficult task of inferring a reward function from other data, such as demonstrations or human preferences over trajectories. In contrast, in our case observed examples of state transitions and rewards from the actual MOMDP can be used directly as training examples to build a reward model via supervised learning.

\subsection{The relationship between reward models and ESR/SER optimization}

\subsubsection{Reward models for SER optimization}
The use of reward models with augmented states has previously been proposed by \citet{vamplew2022impact} in the context of learning SER-optimal policies for environments with stochastic rewards. They argued that for this task the augmented state should be based not on the actual rewards accumulated in the current episode, but on the discounted sum of the expected reward for each state-action-state transition experienced so far in the current episode. This already requires the agent to learn a model of those expected rewards, and use this to derive the augmented state (i.e. $s_t^A = s_t \cup \sum\limits^{t-1}_{i=0} \gamma^i {\bf r^P}_i$). While \citet{vamplew2022impact} only consider the use of the reward model during learning, clearly this model could also continue to be used to provide the required information for the augmented state once the learned policy is deployed.

\subsubsection{Reward models for ESR optimization}
In contrast an ESR-optimal policy is tailored to maximise the utility derived in each individual episode. Therefore where rewards are available, the augmented state should be based on the accumulated actual rewards. Therefore it may seem that the correct option is to train the agent on augmented states derived from the true rewards, and then if true rewards are not available at deployment, apply that same policy but with proxy rewards from the reward model used in place of the actual rewards. However, as noted by \citet{baisero2022asymmetric}, "the optimal action for a partially observable agent can differ greatly from that of an optimal fully observable agent".

\begin{figure}
\centering
\includegraphics[width=.6\columnwidth]{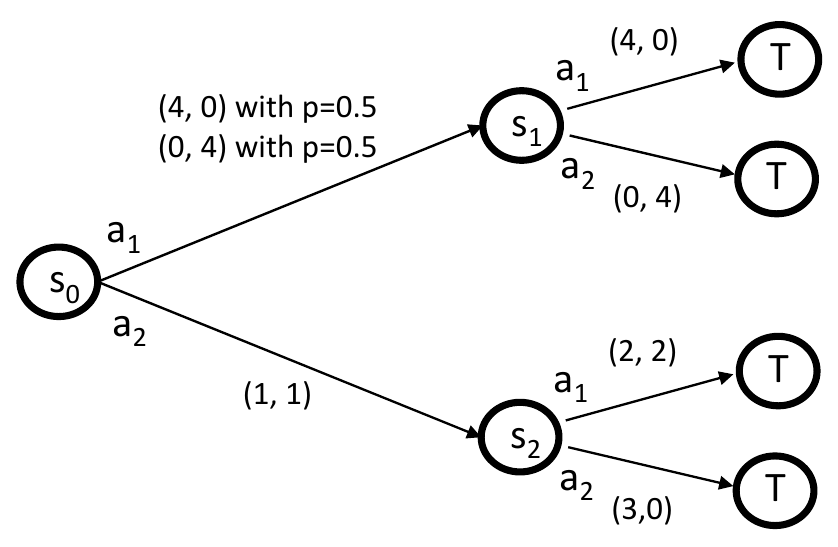}
\caption{\label{fig:3-state-momdp-esr} A simple MOMDP illustrating that the optimal policy may be different for the fully-observable case where the actual reward on each transition is available, and the case where rewards can not be observed and a proxy reward model is used instead.}
\end{figure}

Figure \ref{fig:3-state-momdp-esr} extends the earlier example from Figure \ref{fig:2-state-momdp-esr} to illustrate this for the case of non-observable rewards. Again we assume that the utility function $u$ is taking the minimum value achieved over both objectives. If rewards are fully observable then the optimal policy for this MOMDP and utility function is to execute action $a_1$ in state $s_o$, and then to execute either $a_1$ or $a_2$ in $s_1$ depending on the reward received on the transition between these states -- this will result in an ESR of 4 in all episodes. 

Now consider what happens if the policy learned under these conditions is applied using proxy rewards derived from a reward model trained on this MOMDP. The reward model will have learned an expected reward of $(2,2)$ for $(s_0, a_1$) (the probability-weighted mean of the observed rewards). All other state-action pairs have deterministic rewards and so the reward model will output rewards matching the true rewards. Consider what happens when the agent encounters state $s_1$. The augmented state will contain the accrued proxy reward value of $(2,2)$, rather than the actual reward which would be either $(4,0)$ or $(0,4)$. Two issues arise here. First, the agent has never been trained on the augmented state $(s_1, 2, 2)$ so its policy may fail to generalise to this novel situation. Second, in this specific case, any policy that does not have access to the true reward will inevitably select the wrong action in $s_1$ half of the time, giving rise to a mean ESR of just 2. In contrast if the agent selects action $a_2$ in $s_0$ and $a_1$ in $s_2$, it will always receive a vector return of $(3,3)$, which gives an ESR of 3.

Therefore the optimal policy for the situation where the true rewards are not available differs from the optimal policy where those rewards are observable. Hence, if we know in advance that rewards will not be available post-deployment, then the appropriate course of action is to train a reward model based on the true rewards, and then to use that reward model rather than the true rewards when training the MORL agent.    

\section{Related work}

\citet{eberhard2024actively} note that even in the context of single-objective RL, for many tasks "\emph{the reward function is clearly defined but costly to evaluate}" and so "\emph{reward evaluation becomes the bottleneck in many real-world environments, e.g., in molecular optimization tasks, where computationally demanding simulations or even experiments are required to evaluate states and to quantify rewards}."
They focus on the impact this has on training rather than deployment, and propose to use active learning to replace expensive ground-truth rewards with rewards modelled using neural networks.

The MORL scenario where rewards are available during training but not during deployment has parallels to the problems addressed in the field of partially observable RL with privileged information \cite{wang2023learning,cai2024provable,li2025leveraging,baisero2022asymmetric}. That problem setting assumes that the agent is learning a policy which will be executed within a environment in which the states are only partially-observable, but that access to the ground-truth state information (the \emph{privileged information}) is available during training. The MORL scenario which we are considering can be framed within this setting by treating the accumulated reward-components of the augmented state as privileged information. \citet{cai2024provable} identify two categories of approach for this form of partially observable RL. \emph{Policy distillation} methods first train a policy on the fully-observable environment, and then use this to derive a policy suitable for the partially-observable environment. Alternatively {privileged value learning} methods first learn a value function conditioned on the privileged information, and then use this to learn a policy for the partially-observable setting. Adapting techniques from either of these categories to the MORL setting may be a promising direction for future research.

\section{Conclusion}

Maximising the value of a non-linear utility function within a multi-objective environment (a MOMDP) requires a MORL agent to condition its policy on an augmented state (a concatenation of the state of the environment and a history of prior rewards). This augmented state information is required both during training and, importantly, also after deployment of the agent. This implication of the use of augmented states has not previously been reported in the MORL literature, and may have significant repercussions for application of MORL methods to problem domains where rewards are expensive or unavailable during the deployment phase. We propose that this issue may be addressed by training a reward model which can be used both during learning of a policy, and also to provide proxy rewards post-deployment.

\printcredits

\section*{Declaration of competing interest}
The authors declare that they have no known competing financial interests or personal relationships that could have appeared to influence the work reported in this paper. 

\bibliographystyle{model1-num-names}

\bibliography{cas-refs}


\end{document}